\documentclass{article}
\usepackage{spconf,amsmath,graphicx}

\usepackage{dblfloatfix}
\usepackage{amssymb}
\usepackage{booktabs}
\usepackage{multirow}
\usepackage{multicol}
\usepackage{adjustbox}
\usepackage{subcaption}
\usepackage[ruled]{algorithm2e}
\usepackage[noend]{algpseudocode}
\usepackage{breqn}

\SetKwInput{KwInput}{Input}                
\SetKwInput{KwOutput}{Output}              

\newcount\Comments  
\usepackage{color}
\newcommand{\kibitz}[2]{\ifnum\Comments=1\textcolor{#1}{#2}\fi}

\title{DDI-COCO: A DATASET FOR UNDERSTANDING THE EFFECT OF COLOR CONTRAST IN MACHINE-ASSISTED SKIN DISEASE DETECTION}
\name{Ming-Chang Chiu$^{\star}$~~Yingfei Wang$^{\star}$ \qquad Yen-Ju Kuo$^{\square}$ \qquad Pin-Yu Chen$^{\dagger}$\vspace{-3mm}}

\address{$^{\star}$ University of Southern California $^{\square}$ National Taiwan University Hospital
$^{\dagger}$ IBM Research \\
$^{\star}$Equal Contributions}

\begin{document}
%
\maketitle
%
\begin{abstract}
Skin tone as a demographic bias and inconsistent human labeling poses challenges in dermatology AI. We take another angle to investigate color contrast's impact, beyond skin tones, on malignancy detection in skin disease datasets: We hypothesize that in addition to skin tones, the color difference between the lesion area and skin also plays a role in malignancy detection performance of dermatology AI models. To study this, we first propose a robust labeling method to quantify color contrast scores of each image and validate our method by showing small labeling variations. More importantly, applying our method to \textit{the only} diverse-skin tone and pathologically-confirmed skin disease dataset DDI, yields \textbf{DDI-CoCo Dataset}\footnote{https://github.com/charismaticchiu/DDI-CoCo}, and we observe a performance gap between the high and low color difference groups. This disparity remains consistent across various state-of-the-art (SoTA) image classification models, which supports our hypothesis. Furthermore, we study the interaction between skin tone and color difference effects and suggest that color difference can be an additional reason behind model performance bias between skin tones. 
Our work provides a complementary angle to dermatology AI for improving skin disease detection.
\end{abstract}
\vspace{-2mm}
\section{Introduction}
\vspace{-2mm}
Recent advances in computer vision have made significant progress in aiding automated diagnostics and decision-making in dermatology AI \cite{tschandl2020human,esteva2017dermatologist}. This field aims to reduce barriers to dermatology care, which could benefit billions of people \cite{coustasse2019use}. However, recent studies have revealed biased performances among different skin tones and lesion types \cite{groh2021evaluating,Daneshjou2022DisparitiesID,daneshjou2021lack,Kinyanjui2020Fairness}, calling for more development in this research area. In addition, it is crucial to recognize that even though individuals with darker skin tones or people of color (PoC) have a lower risk of developing skin cancer, they face a higher mortality rate from the disease, primarily because of delays in detection or seeking medical attention \cite{gupta2016skin}. Therefore, we aim to provide more insights for evaluating darkly pigmented lesions in PoC by our technique.
Our work takes a different angle and asks whether, in addition to skin tone, the color contrast between the lesion and the background could be a factor that also affects the prediction of dermatology AI. 

\begin{figure}[t]
    \centering
    \includegraphics[width=1\linewidth]{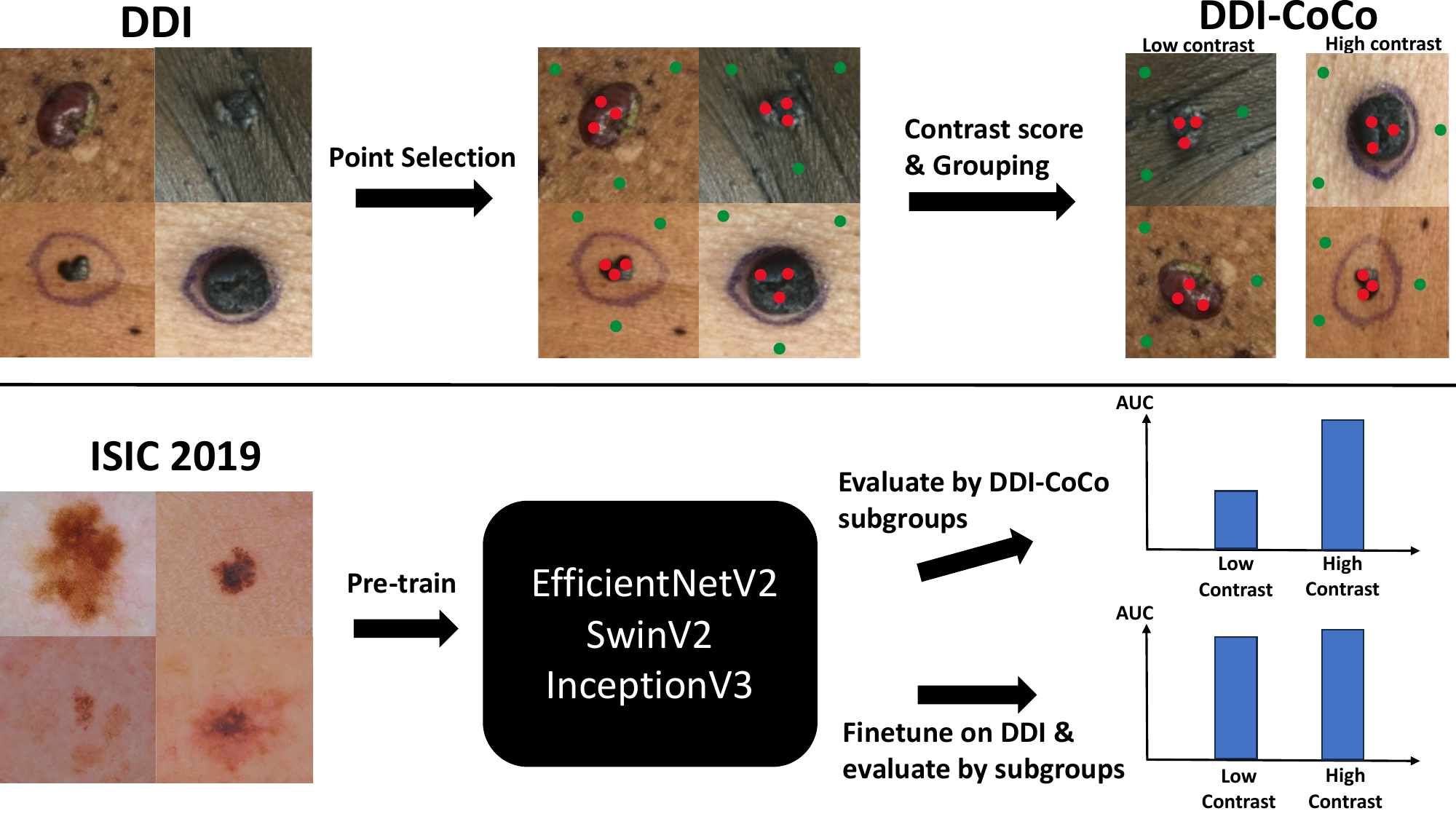}\vspace{-3mm}
    \caption{\textbf{Labeling process \& experimental setups.} 
    }
    \label{fig:label_flow}
    \vspace{-5mm}
\end{figure}

We introduce a novel \textit{dermatologist-reviewed and variability-controlled labeling technique} for the quantitative assessment of contrast scores in clinical images. Our aim is to establish a bias-mitigating labeling procedure that offers a cost-effective approach by involving advanced tests only when necessary. This approach is driven by the inadequacies in achieving a consensus among dermatologists, as highlighted in recent studies \cite{Daneshjou2022DisparitiesID}. We will validate the robustness of our labeling process to human factor in Sec.~\ref{sec:methods}.

Next, we apply our dermatologist-reviewed labeling method to the recently released Diverse Dermatology Images (DDI) dataset \cite{Daneshjou2022DisparitiesID}, yielding our proposed \textbf{DDI-Co}lor \textbf{Co}ntrast (\textbf{DDI-CoCo}) dataset that has contrast scoring on each image. We choose the DDI dataset as our test bed since it serves as \textit{the only} dermatologist-annotated and pathologically proven clinical image dataset with diverse skin tones and diverse diseases, and is shown to reduce skin tone bias. 

We systematically assess color contrast's impact on malignancy detection in skin diseases using three deep neural network (DNN) models on DDI-CoCo dataset. Our approach serves a twofold purpose: first, we evaluate the performance of image classification models trained on the commonly-used ISIC 2019 dataset, and second, we investigate their performance after fine-tuning on our DDI-CoCo dataset. 

We emphasize the development of a dermatologist-supervised labeling framework and its importance in enabling the study of color contrast effects in skin disease prediction. We also highlight the robustness of our labeling technique, which helps to address biases and offers valuable insights for improved understanding and diagnosis of skin diseases.

\vspace{-5mm}
\section{Related Work}\label{sec:related}
\vspace{-2mm}
 Research has shown superior results from DNNs \cite{esteva2017dermatologist, Cassidy2021AnalysisOT} than experts in clinical skin disease datasets, and the International Skin Imaging Collaboration (ISIC) has also organized valuable annotated datasets for this domain. 
 Prior works in dermatology AI have tested with setups like using InceptionNetV3 (DeepDerm\footnote{One of our goals is to fairly compare SoTA architectures with InceptionNet V3, so we avoid using pre-computed DeepDerm which trains on proprietary datasets.})\cite{Daneshjou2022DisparitiesID,esteva2017dermatologist} to train on \textit{proprietary} large-scale datasets, implementing small and computationally efficient models such as MobileNet \cite{chaturvedi2021skin} or modifying VGG-16 \cite{groh2021evaluating} for skin-lesion classification, or studying fair classification using automatic skin tone scoring using DenseNet \cite{Kinyanjui2020Fairness}. 
 
 Many prior works have relied on visual consensus labels from a small group of dermatologists \cite{esteva2017dermatologist,daneshjou2021lack,liu2020deep}. However, this process can be noisy as they may lack information useful for diagnoses such as clinical history, in-person visit review, and other diagnostic tests. \cite{hekler2020effects} showed that a melanoma classifier trained on dermatologists' consensus dataset performed significantly worse than models trained on pathologically confirmed ground truth dataset. This motivates us to design a labeling method that not only shows clinical value but also has a low human factor effect.
 
 On the other hand, color contrast has been an aspect used by Ophthalmologist to test the vision capability of human beings \cite{biochem_texbook,neurosci_book,Chiu2022OnHV} and \cite{lester2021clinical} indicates biases in photography such as color balancing could contribute to the difficulty of capturing critical features. Classical computer-aided diagnosis methods often rely on image pre-processing or curating color-based features to assist classification or segmentation \cite{chen2003colour, cheng2008skin,madooei2016incorporating}. Recent work on deep learning based skin lesion detection also begin to incorporate color and illumination factors such as color bands, shading attenuation, and greyscale transformation \cite{abhishek2020illumination}, demonstrating the applicability of color theory to modern deep learning based frameworks. However, incorporating these factors as high dimensional features limits model interpretability, and fails to allow for direct interrogation of intrinsic characteristics presented in lesion-skin color contrast, and also, \cite{Chiu2022OnHV, Chiu_2023_ICCV} suggests DNNs has built-in color and contrast bias. Thus, we aim to study how lesion-skin color contrast plays a role in dermatology AI and specifically in malignancy detection. We highlight that the skin-lesion contrast in our work is distinct from the overall image contrast as measured by illumination or color theory, as adjusting the overall image contrast will not necessarily lead to a mitigation of skin-lesion contrast bias. Note that we use color contrast and color difference interchangeably in this work.

\begin{figure}[htb!]
    \begin{minipage}[b]{0.48\linewidth}
         \centering
         \includegraphics[width=\textwidth]{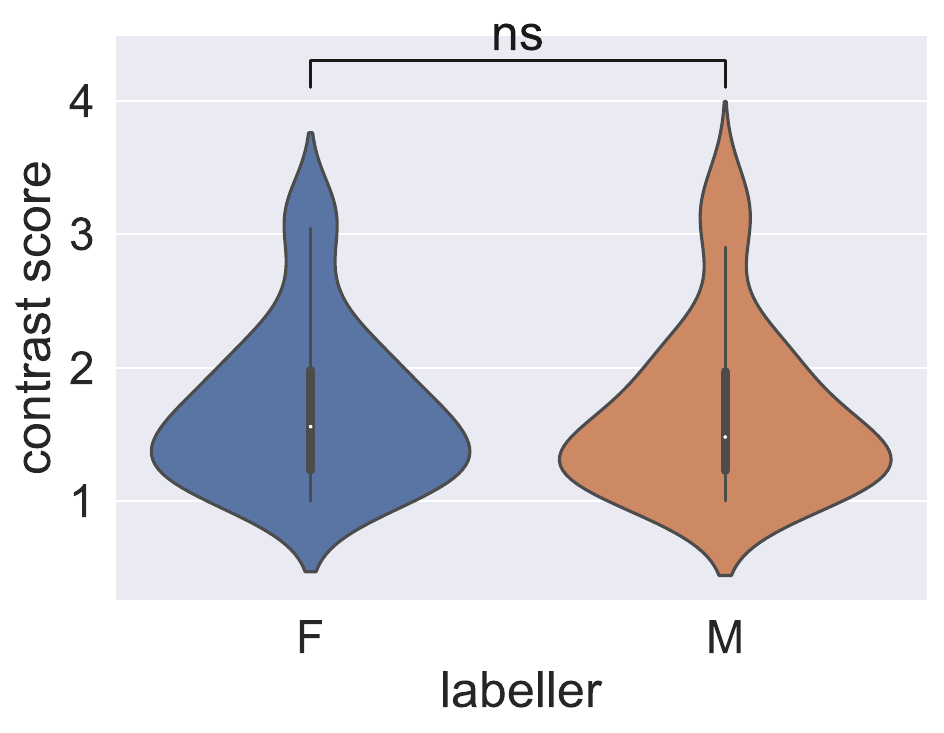}\vspace{-2mm}
         \centerline{(a)}
    \end{minipage}
    \begin{minipage}[b]{0.49\linewidth}
     \centering
         \includegraphics[width=\textwidth]{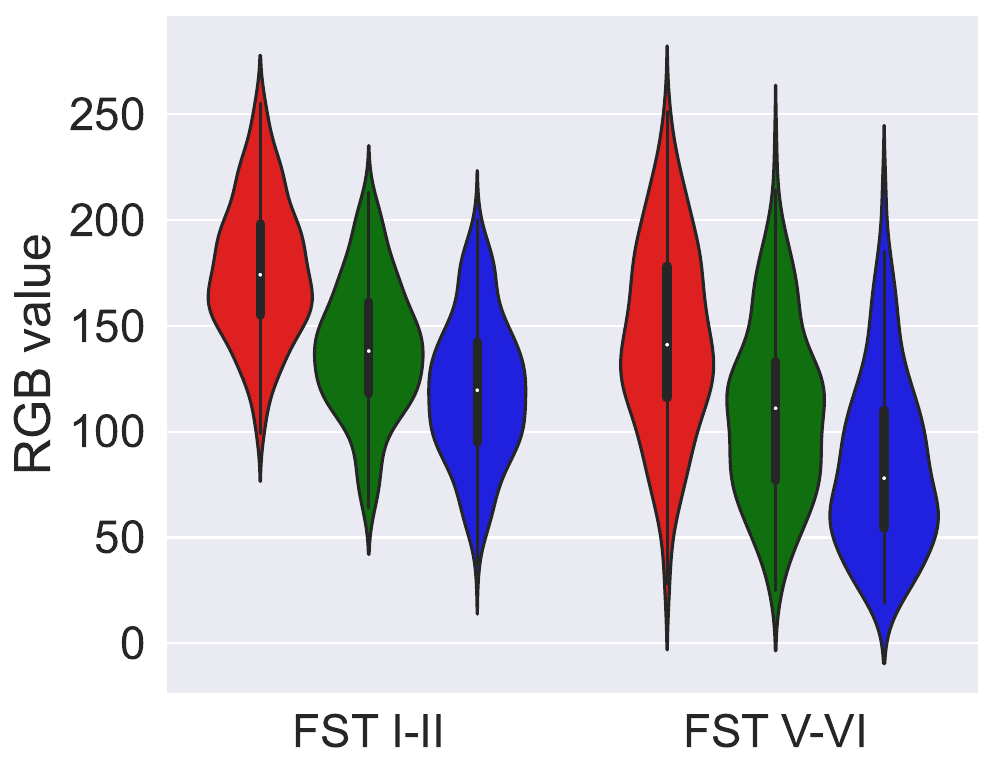} 
        \centerline{(b)}
    \end{minipage}\vspace{-3mm}
    \caption{\textbf{(a) Distributions of contrast scores by labeled. (b) Background RGB by FST skin tones. Contrast scores show high consistency between the two labellers, and background point RGB values correlate with the dermatologists-confirmed FST skin tone groups.}
    }
    \label{fig:justifications}
    \vspace{-6mm}
\end{figure}
 \vspace{-5mm}
\section{Datasets}\label{sec:datasets}
\vspace{-2mm}
  To enable fair comparisons of different models on DDI-CoCo, we follow common practice and adopt the publicly available ISIC 2019 as the base training dataset. ISIC 2019 is the most commonly used dataset for AI algorithm development in skin disease classification, and it also includes various patient metadata for potential deeper analysis. We do not possess the right to share the DDI dataset, but it is publicly available and we will share our DDI-CoCo. 
  \vspace{-5mm}
  \subsection{ISIC 2019 Dataset}
  \vspace{-2mm}
  ISIC 2019 \cite{combalia2019bcn20000,Tschandl2018_HAM10000,codella2018skin} contains 25,331 images and includes the whole popular HAM10000 (ISIC2018 training set) in addition to the lesions in hard-to-diagnose locations such as nails and mucosa. 
  \vspace{-5mm}
  \subsection{Diverse Dermatology Images (DDI) Dataset} 
  \vspace{-2mm}
  The DDI dataset contains 656 images from 78 skin diseases and is \textit{the first and only} pathologically confirmed dataset with diverse Fitzpatrick skin types (FST), a clinical classification scheme for skin tone. It has relatively balanced sizes across three different skin tone groups, 208 images of FST I-II (light skin tones), 241 images of FST III-IV, and 207 images of FST V-VI (dark skin tones), and all skin tones are determined using in-person visits reviews by two board-certified dermatologists. 
  Note that DDI was \textit{originally proposed to compare FST V-VI group and FST I-II group}, as a result, we will also \textit{only} consider these two groups in the analysis. 
  
\vspace{-5mm}
\section{Methods}\label{sec:methods}
\vspace{-2mm}
  In this section, we present our dermatologist-reviewed labeling and contrast scoring process, which involves annotating the contrast score for each image. Also, we provide an overview of the experimental setups employed in our study.

\vspace{-5mm}
  \subsection{Labeling Process}
  \vspace{-2mm}
   We apply a mathematical derivation based on the average pixel color intensities to annotate each image with a contrast score reflecting the contrast between the lesion and skin tone. This approach ensures robustness against human labeling bias and provides objective measurements for quantifying the contrast. 
   We justify our method in Sec.~\ref{sec:label} and detail our labeling process as follows:
    \vspace{-2mm}
    \begin{enumerate}
        \item We manually retrieve the RGB values of three random points on the lesion area (within the circular diagnostic markings when presented). We avoid choosing points on active bleeding site of ulcerative lesions, dark-colored scar tissue regions or any consequences interfering the color of primary lesions. We name these points foreground points (red dots in Fig \ref{fig:label_flow}). 
        \vspace{-2mm}
        \item We retrieve RGB values of three random points around the lesion area. We retrieve points on normal perilesional skin, which is the same color as the patient's skin tone, without any underlying erythema, nor fibrosis associated with possible inflammatory change. To decrease the risk of surrounding inflammation-induced color changes confounding the primary lesion color, we avoid choosing points immediately adjacent to the lesion. We name these points background points (green dots in Fig.~\ref{fig:label_flow}).
        \vspace{-2mm}
        \item When the lesion lies in an area with a special lighting condition, such as in the shadow, the labellers are instructed to choose foreground and background points in similar lightning conditions. 
        \vspace{-2mm}
        \item We average the three RGB points for background and foreground, respectively, and compute the contrast score between the average foreground and background colors with the official WCAG formula \cite{w3c2008WCAG} for accessibility, emulating the idea from \cite{Chiu2022OnHV}.
    \end{enumerate}

\vspace{-4mm}
 \subsection{Experimental Setups} 
 \vspace{-2mm}
 We use Inception V3 for its commonality and explore the SoTA models such as the EfficientNetV2-S (more efficient) and Swin Transformer V2-B (bigger). All models are pre-trained on ImageNet. The objective of our evaluation is twofold: (1) to evaluate if SoTA models are also subject to contrast bias, similar to the skin tone bias observed in the DDI paper, and (2) to assess the efficacy of DDI-CoCo in reducing the performance gap between high and low contrast groups. For the first goal, we train on the comprehensive and publicly available ISIC 2019 dataset and evaluate on the entire DDI-CoCo dataset. This out-of-distribution (OOD) setup is practical as we do not always have in-distribution data in clinical evaluations. To achieve the second goal, instead of evaluating the entire DDI-CoCo, we fine-tune on 80\% of images and evaluate on the rest and we follow fine-tuning process in \cite{dosovitskiy2020vit}. 

 All experiments and data splits are performed with \textbf{five different random seeds}. The results of the experimental replicates are shown as error bars in all the model performance plots. We use weighted cross-entropy loss to account for any class imbalance, and the weights are obtained from the inverse frequency of each class.
 
 We filter out 14 images due to their borderline pixel intensities causing abnormal contrast scores per WCAG formula, and then divide the rest of the images in DDI-CoCo into two contrast groups, ``high" and ``low" (see example images in Fig.~\ref{fig:label_flow}), by a cutoff at the median of contrast scores. Tab.~\ref{tab:contrast} summarizes the statistics of DDI-CoCo, a balanced high/low contrast presence among skin tones was achieved.
 
\begin{figure*}[htb!]
    \begin{minipage}[b]{0.33\textwidth}
         \centering
         \includegraphics[width=\linewidth]{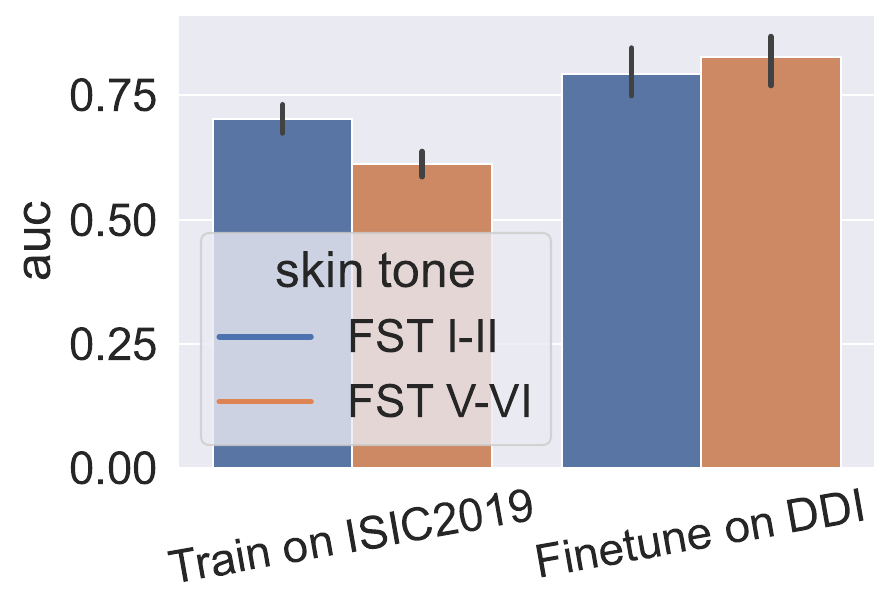}\vspace{-2mm}
         \centerline{(a) Replicating DDI result on InceptionV3}
     \end{minipage}
     \begin{minipage}[b]{0.33\textwidth}
         \centering
         \includegraphics[width=\linewidth]{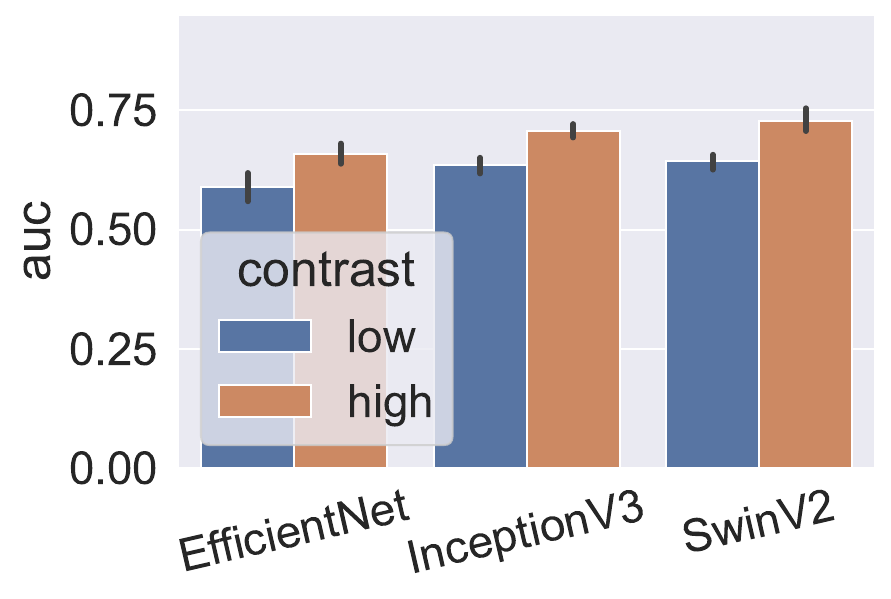}\vspace{-2mm}
         \centerline{(b) color contrast in effect}
     \end{minipage}
     \begin{minipage}[b]{0.33\textwidth}
         \centering
         \includegraphics[width=\linewidth]{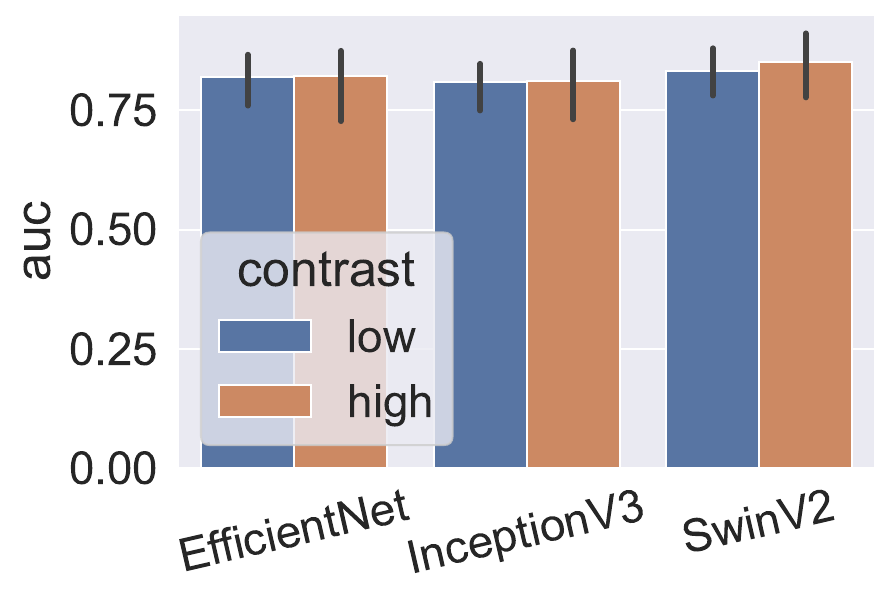}\vspace{-2mm}
         \centerline{(c) fine-tuning in effect}
     \end{minipage}\vspace{-2mm}
    \caption{\textbf{(a) We replicate the results of DDI paper by showing performance bias in different skin tone groups, and by showing fine-tuning reduces this performance gap. (b) DNNs perform consistently better in high color contrast group, showing the bias caused by lesion-skintone color contrast. (c) fine-tuning on DDI-CoCo reduces the performance gaps.}}
    \label{fig:replicate_n_contrast}
    \vspace{-4mm}
\end{figure*}
\begin{figure*}[htb!]
    \begin{minipage}[b]{0.24\textwidth}
         \centering
         \includegraphics[width=\linewidth]{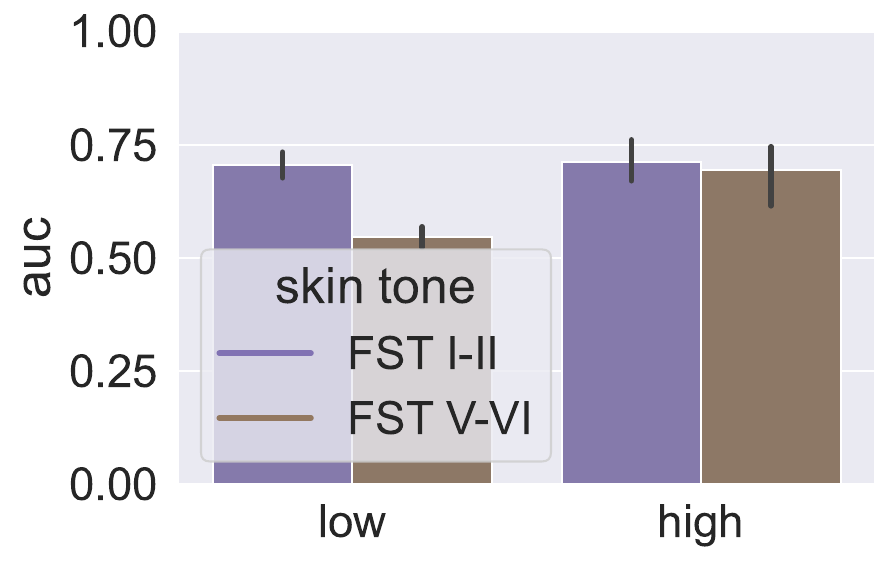}\vspace{-2mm}
         \centerline{(a)}
     \end{minipage}
    \begin{minipage}[b]{0.24\textwidth}
         \centering
         \includegraphics[width=\linewidth]{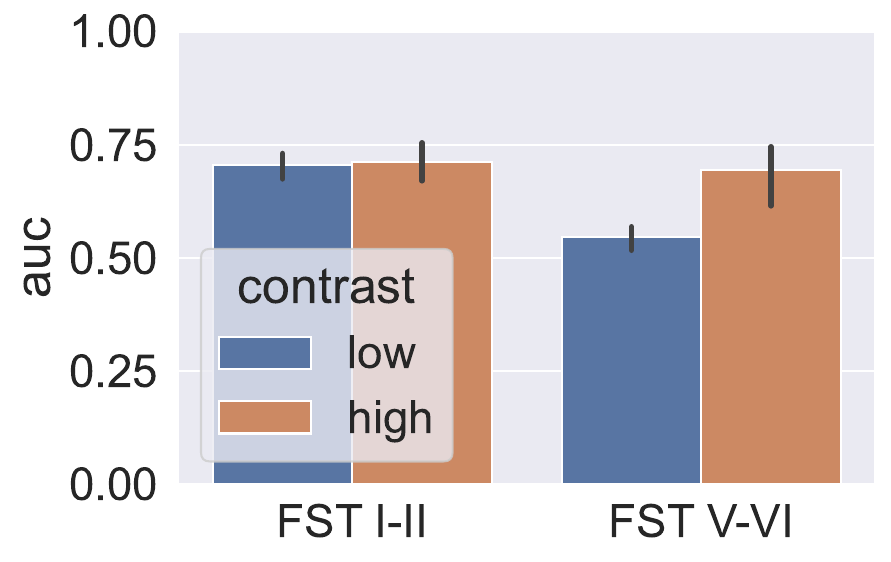}\vspace{-2mm}
         \centerline{(b)}
     \end{minipage}
     \begin{minipage}[b]{0.24\textwidth}
         \centering
    
    \includegraphics[width=\linewidth]{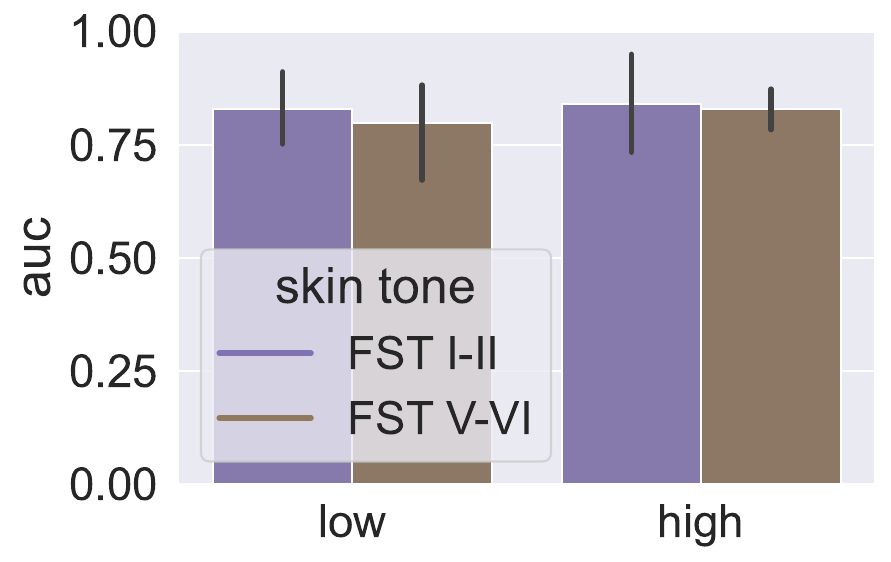}\vspace{-2mm}
         \centerline{(c)}
     \end{minipage}
     \begin{minipage}[b]{0.24\textwidth}
         \centering
         \includegraphics[width=\linewidth]{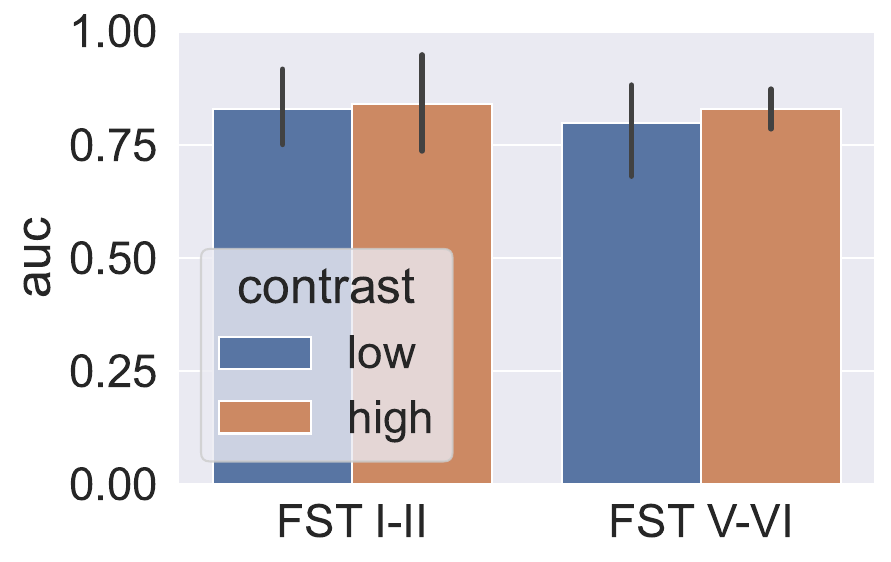}\vspace{-2mm}
         \centerline{(d)}
     \end{minipage}\vspace{-2mm}
    \caption{\textbf{(a) AUC by contrast group before fine-tuning. (b) AUC by skin tones before fine-tuning. (c) AUC by contrast group after fine-tuning. (d) AUC by skin tone after fine-tuning} 
    }
    \label{fig:contrast_n_skintone}
    \vspace{-5mm}
\end{figure*}

\vspace{-5mm}
\section{Results and Analysis}
\vspace{-4mm}
In this section, we first demonstrate the robustness of our method by analyzing point selections and showcasing the high scoring consistency achieved between different labellers, then validate our twofold evaluations by comparing InceptionNet V3 (DeepDerm) performance on different skin tones. The skin tone effect result indicates that our experimental setup is able to replicate the result in the original DDI paper (Fig.~\ref{fig:replicate_n_contrast}(a)), which corroborates the validity of our experimental setups. Next, we discuss the effect of lesion-skin tone contrast on malignancy detection. Furthermore, we dive deeper into each skin tone type to discuss and explore the interaction of contrast and skin tone. Lastly, we discuss how SoTA models' behaviors. 

\begin{table}[t]
    \centering
    \begin{tabular}{lcccc}\toprule
     DDI-CoCo    &  I-II & III-IV & V-VI & Total\\\midrule
    High & 88    & 134 & 92 & 314 \\
    Low  & 114   & 103 & 111 & 328 \\\bottomrule
    \end{tabular}\vspace{-3mm}
    \caption{\textbf{Number of images by contrast \& skin tones.}}
    \label{tab:contrast}
    \vspace{-4mm}
\end{table}


    

\vspace{-5mm}
\subsection{Justification of Labeling Process}\label{sec:label}
\vspace{-2mm}
 We justify our labeling approach and demonstrate contrast score validity by showing high between-labeller consistency. Fig.~\ref{fig:justifications}(a) shows that the contrast score distributions are similar across two labellers trained by a dermatologist (F and M), and a paired t-test with a p-value of 0.95 further validates the identical contrast score distribution. 
 These results show our designed labeling process is robust to human factors, and can reliably generate contrast scores, despite the randomness in the initial foreground and background point selection processes.
 In addition, for background points that are chosen to capture skin tone colors, our labeling method shows a strong correlation between RGB values and skin tone, where darker skin tone (FST V-VI) has smaller RGB values (Fig.~\ref{fig:justifications}(b)), indicating the validity of background point selection.

 This consistency is encouraging as it showcases we can apply our process as pretest in areas with scarce medical resources.

  \vspace{-6mm}
  \subsection{Color Contrast in Effect}
  \vspace{-2mm}
    In the initial phase of our dual evaluation, where we train models on ISIC 2019 and then assess their performance on the complete DDI-CoCo dataset, we observe a performance bias favoring images with high lesion-skintone color contrast. Fig.~\ref{fig:replicate_n_contrast}(b) shows the tested models' AUC of the high and low contrast groups. This finding underscores the vulnerability of dermatology AI models to bias associated with lesion-skin tone contrast when trained on clinical images. 
  
  \vspace{-4mm}
  \subsection{Fine-tuning closes the performance gap between low and high contrast groups}\vspace{-2mm}
    In pursuit of our second evaluation objective - fine-tuning, as depicted in Fig.~\ref{fig:replicate_n_contrast}(c), we observe a more equitable performance distribution. This underscores two key points: (1) Color contrast, similar to skin tone, can introduce OOD issues when using dermatology AI algorithms without diverse representation in the training dataset, and (2) Fine-tuning on a diverse dataset effectively narrows the performance gap between high and low contrast groups.
    
  \subsection{Diving Deeper into Skin Tones}
    The DDI paper shows that skin tone can be a major attribute behind dermatology AI's performance bias. However, we believe the unbalanced skin tone representation in the training set only partially explains why such a bias arises. Therefore, we propose that lesion-skin tone contrast is another complementary underlying reason, especially since the majority of lesions appears in dark color. Fig.~\ref{fig:contrast_n_skintone}(a) pinpoints the significant skin tone effect in low-contrast images, but nearly no difference can be observed in the high contrast group. Similarly, Fig.~\ref{fig:contrast_n_skintone}(b) shows that the contrast effect is more significant in the dark skin tones (FST V-VI). After fine-tuning on DDI-CoCo, both effects are mitigated (Fig.~\ref{fig:contrast_n_skintone}(c) \& (d)). 
    \vspace{-2mm}
  \subsection{Similar Behavior in SoTA Model}
 A surprising discovery in our study is that SoTA models exhibit performance patterns similar to Inception V3 (Fig.~\ref{fig:replicate_n_contrast}(b) \& (c)), despite that ViT \cite{dosovitskiy2020vit} claims to possess less inductive bias. Across various model architectures, we consistently observe superior performance in the high-contrast group, echoing the findings in \cite{Chiu2022OnHV}. Initially, there is a substantial performance gap between the high and low contrast groups, which diminishes after fine-tuning. This unexpected consistency in trend challenges our expectations regarding the behavior of SoTA models, suggesting that fine-tuning plays a crucial role in closing performance gaps across different model architectures.
  \vspace{-1mm}

  \begin{table}[t]
    \centering
    \begin{tabular}{cccc}\toprule
     AUC &  Inception & Efficient & SwinV2\\
     \midrule
    High & 0.810 & 0.822 & 0.852  \\
    Low & 0.809 & 0.819 & 0.832 \\
    \bottomrule
    \end{tabular}\vspace{-2mm}
    \caption{\textbf{Average AUC by contrast after fine-tuning.} 
    }
    \label{tab:performances_contrast}
    \vspace{-3mm}
\end{table}

\begin{table}[t]
    \centering
    \begin{tabular}{cccc}\toprule
     AUC &  InceptionV3 & EfficientNet & SwinV2\\
     \midrule
    I-II & 0.792 & 0.776 & 0.814  \\
    V-VI & 0.828 & 0.862 & 0.895 \\
    Gap & +0.036 & +0.086 & +0.081\\
    \bottomrule
    \end{tabular}\vspace{-2mm}
    \caption{\textbf{Average AUC by skin tones after fine-tuning.} 
    }
    \label{tab:performances_skintone}
    \vspace{-5mm}
\end{table}

\vspace{-3mm}
\section{Conclusion}
\vspace{-3mm}

Our systematic investigation focused on the influence of color contrast on skin disease malignancy detection. We introduced the DDI-CoCo dataset, incorporating a dermatologist-reviewed contrast labeling method to mitigate human labeling variations. This labeling method holds promise for application in resource-constrained settings. Our results highlight the significant impact of contrast on skin disease detection, with the high-contrast group consistently outperforming the low-contrast group. Additionally, we found that fine-tuning models on diverse datasets effectively mitigated both contrast and skin-tone performance biases, underscoring the complimentariness of our approach.





\bibliographystyle{IEEEbib}
{\small
\bibliography{strings,refs}
}
\end{document}